# Multi-Topic Multi-Document Summarizer

Fatma El-Ghannam[1] and Tarek El-Shishtawy[2]

[1] Electronics Research Institute, Cairo, Egypt
[2] Faculty of Computers and Information, Benha University, Benha, Egypt

## ABSTRACT

*Current multi-document summarization systems can successfully extract summary sentences, however with many limitations including:  low coverage, inaccurate extraction to important sentences, redundancy and poor coherence among the selected sentences. The present study introduces a new concept of centroid approach and reports new techniques for extracting summary sentences for multi-document. In both techniques keyphrases are used to weigh sentences and documents.   The first summarization technique (Sen-Rich) prefers maximum richness sentences. While the second (Doc-Rich), prefers sentences from centroid document. To demonstrate the new summarization system application to extract summaries of Arabic documents we performed two experiments.   First, we applied Rouge measure to compare the new techniques among systems presented at TAC2011. The results show that Sen-Rich outperformed all systems in ROUGE-S. Second, the system was applied to summarize multi-topic documents. Using human evaluators, the results show that Doc-Rich is the superior, where summary sentences characterized by extra coverage and more cohesion.*

## KEYWORDS

*1. Summarization,   Multi-document summarization,   keyphrase-based summarizatio,  keyphrase extraction,    Topic identification,    information retrieval.*

## 1. INTRODUCTION

With the continuing growth of online information, it has become increasingly important to provide improved mechanisms to find and present a summary of textual information, not only for a single document but also for multiple documents. Multi-document summarization is an automatic procedure that extracts important information from multiple documents. Many efforts have focused on extracting a representative summary either from single or multiple documents. Single document summarization is a difficult task by itself, but Multi-document summarization (MDS) has additional challenges.

The major challenge of MDS is due to the multiple resources from which information is extracted. Multiple documents include the risk of higher redundant information than would typically be found in a single document. In addition, the ordering of extracted information from a set of documents into a coherent text in order to create a coherent summary is a non-trivial task [3].

Summarization can be either extractive or abstractive. Extractive summarization involves assigning saliency measure to some units (e.g. sentences, paragraphs) of the documents and extracting those with highest scores to include in the summary. Abstractive summarization usually needs information fusion, sentence compression and reformulation. Abstractive summarization is a difficult problem because it requires deeper analysis of source documents and concept-to-text generation. Currently most of the researches and commercial systems in automatic text summarization are extractive summarization. Concerning generality of summaries, two types can be distinguished: generic and query-driven summaries. The first type tries to represent all relevant topics of a source text, while the second focuses on the user's desired query keywords or topics.

Most of the existing successful summarization systems are used in domain of news articles where each document is assumed to have a 'mono-concept'. It is assumed in these systems that a document has information about a single event, accident, or news. In such systems, one of the key tasks is to cluster multiple documents either on time bases or on topics extracted from user-input query. For example MEAD [17] selects centroid sentence of each cluster, and searches for similar or strongly related sentences to centroids. CLASSY (Schlesinger et al., 2008) ranks sentences with their inclusion of user query terms and their associated signature words.

In the proposed work, a new concept-centroid approach is presented for multi-document summarization. It fits the following scenario: "User is faced with a collection of related documents which have information about a main subject. The subject has different topics which are unknown to user. Further, each document discusses one or more topics of the target subject. User wishes to have summary highlighting important information contained in the collection. In multi-topic documents, classifying documents at earlier stages of summarization process may lose important topics covered. Also, making use of user query terms (even with their linguistically equivalent words) may not be suitable for users who don't know all aspects of a certain topic.
The system automatically extracts the key shared concepts and their relevant information among all the documents plus other information unique to individual documents. Extracted keyphrases are used to evaluate both the concept richness of each document, and the importance of each sentence in the cluster. A new scoring scheme for keyphrases extracted from multiple documents is presented. Two keyphrase based techniques for extracting summary sentences are presented.
The presented work is implemented to extract summaries of Arabic documents; however, the approach is language-independent. To the best of our knowledge, this is the first work that exploits automatically extracted keyphrases to produce Arabic multiple document summaries. The system is characterized by:

- The presented approach is a generic extractive summary for multi-topic, multi-document summarization.
- The decision of extracting sentences to be included in the summary is postponed until evaluation of the whole picture of topics in all documents.
- Domain independent topics are extracted automatically, and the system ranks documents and sentences by their topic richness. Centroid documents containing highly ranked sentences are then identified.
- The system balances between concept importance and coverage of all topics.
- Flow of summarized texts follows centroid documents and is augmented by sentences extracted from other documents.

The remaining of this paper are organized as follows: Section 2 reviews the previous works; Section 3 the proposed system; Section 4 discusses evaluation techniques and results; and section 5 is the conclusion.

## 2. PREVIOUS WORKS

A variety of summarization methods have been developed recently. Extractive methods share common tasks to generate an effective summarization. In the following sections we describe basic processes:

**Coverage:** Extraction process plays a central role in summarization process, as it identifies important information that covers different topics in source documents. Extraction process can be performed at any level of textual passages: paragraph, sentence, phrase, word, etc. Various algorithms have been proposed to identify salient information from the source documents [11]. Most of these methods first identify important words or phrases from the source documents using term-weighting methods such as TF-IDF [22], and then extract passages that contain these words or phrases. Many extractive summarization systems consider sentences selection as the final goal.

**Coherency:** Optimal ordering of selected sentences to create a coherent context sequence flow is a difficult problem. In single document summarization, one possible ordering of extracted sentences is provided by the input document itself [12]. However, ordering a set of sentences extracted from a set of documents into a coherent text is a non-trivial task (Bollegala, 2007). In general, methods for sentence ordering in MDS can be classified into two approaches: making use of chronological information [13;16] and learning the natural order of sentences from large corpora.

**Redundancy elimination**: Due to length limitations required for an effective summary, and the existence of many extracted sentences that include the same information; it is desirable to select just one of them to include in the summary. Many researchers [1; 10] use the similarity measure in different ways to identify the duplicate information.

### Existing MDS systems:

Five existing MDS systems which were presented as multiling pilot at TAC 2011 [7] were used in our experiments. We reviewed these systems in addition to other systems related to the same domain.

**MEAD** [17] is a news tracking and summarizing system based on a centroid-based approach [18]. It works by clustering documents describing an event. Clusters are chosen to include two or more documents ordered by time. The system then constructs cluster centroids consisting of words best describing that cluster. Each sentence within the cluster is scored based on its similarity of centroid words, and n sentences are selected to best represent the cluster. The process is repeated for each cluster, and the extracted sentences are ordered based on news time stamps to constitute the summary. In other work, MEAD was extended [19] to allow the system to collect clusters either by user input keywords, or keywords extracted from example news. El_Haj et. al. [4] extended the MEAD concept to include the Arabic language. They used different parameter settings for extracting sentences including the cluster size and the selection model.

**CIASSY** [23] is an event-focused, query-based multi-document summarization. It summarizes Arabic documents by translating documents into English to extract the summary. The system uses language dependent trimming rules to focus on important parts of the sentences. Sentences are then ranked by their inclusion of "signature words"- which occurs significantly more than expected in the document.

**MMR-MD** [8] is a query-relevant MDS, in which each passage (sentence) is ranked by summing its cosine similarity to user input query, its similarity to other passages within cluster, its inclusion of specific word types (such as named entities) and its time stamp sequence; the latest of which is preferred. To achieve diversity, the ranking score of a passage is penalized by its cosine similarity to previously selected passages. It is penalized also if it is part of previously selected clusters and documents.

**Lexical Chain** method [2] disambiguates word senses using shallow syntactic analysis, part-of-speech tagger, and WorldNet thesaurus. The system organizes semantically related noun words as lexical chains. Sentences corresponding to the strongest lexical chains are then extracted to construct the summary.

**CIST**: Liu et. al. [15] introduces an extractive multi-document summarization method for 7 languages including Arabic. The method constructs a hierarchical tree structure of candidate sentences based on hierarchical topic model of hierarchical Latent Dirichlet Allocation (hLDA). Each sentence is represented by a path in the tree, and each path can be shared by many sentences. The assumption is that sentences sharing the same path should be more similar to each other because they share the same topics. Also, if a path includes a title sentence, then candidate sentences on that path are more likely to be included in the generated summary. The system extracts the sentences with the highest scores to include into the summary. Features for sentences assessment are (1) the similarity of title sentences with all candidate sentences; (2) the number of sentences that are assigned on each path, assuming the path that contains more sentences is considered to be the main topic or hot topic; (3) the number of named entity that one sentence contains; and (4) word frequency in sentences. To remove duplicate sentences, similarity is calculated considering the number of intersecting words between the two sentences. If the similarity between a candidate sentence and the sentences that have been added to the summary exceeds a certain threshold, the present candidate sentence is ignored and the system re-selects another sentence.

The system by Saggion [20] TLAN_UPF introduced a technology to produce multiple document generic summaries in four languages including Arabic, based on a single unsupervised system to deal with the four languages. Vector representations are created for each sentence and each document based on statistics of word frequency. The centroid of the cluster is computed as the average of all document vectors. Sentence ranking is based on the similarity between the sentence vector and the centroid vector using the cosine measure. This method differs from the cluster centriod method used in Mead where centroids are the top ranking tf-idf sentences. For sentences belonging to the same document, ordering follows the original document, while documents are sorted according to their document ID in the dataset. This ordering technique differs from system [4] where ranking of sentences in summary is done according to similarity to the centroid.

Most of the methods that depend heavily on sentence similarity and more particularly those related to multi-lingual summarization, focus on the frequency of reoccurring terms and approach the summarization problem as a bag of words, ignoring the dependency of lexical and syntactic features in the extracted summary. This is one of the reasons that motivated us to exploit the existing Lemma Based Arabic Keyphrase Extractor LBAKE [6] module - a module based on both statistical and linguistic features - as a starting point for the proposed multiple documents summarization. The system automatically extracts the keyphrases for individual documents. Extracted keyphrases are used to evaluate both the concept richness of each document, and the importance of each sentence in the cluster. A new cluster topic scoring technique for the keyphrases extracted from multiple documents is presented. In this technique the main concepts are represented by keyphrases that maximize both the relevance and coverage to the cluster. We explore balances between concept importance and coverage of all topics. Sentences and documents are assessed based on the global keyphrase importance. The system ranks documents and sentences by their topic richness. Centroid documents are then found which contain highly ranked sentences. Ordering of the summarized texts follows the centroid document and is augmented by sentences extracted from other important documents.

# 3. THE PROPOSED SYSTEM

Generating an effective extractive summarization system requires many processes. A central process is the assessment and selection of pieces of information according to their relevance to a particular subject or purpose. Instead of using text passages (documents, paragraphs, or sentences) directly for assessment, the current work relies on extracting keyphrases which are used as attributes to evaluate the topic's richness of sentences and documents within the cluster. The steps of the summarization algorithm are similar for both the Sen-Rich and the Doc-Rich technique; however step 4, as noted below, utilizes a different approach for Sen-Rich than the approach for Doc-Rich. The steps are:

1.  Extract local keyphrases for each document.
2.  Construct cluster topics of the set of documents.
3.  Assess the cluster sentences and documents.
4.  Extract important sentences to construct the summary.

In the first step, keyphrases contained at each document were extracted separately. Each keyphrase was assigned a local keyphrase score based on its importance in the document. Then, each keyphrase was assigned a new global cluster topic score based on its importance on the local document as well as the relevance to the cluster. In the third step, all sentences of the documents were assessed based on their richness of important cluster topics. The fourth step involved analyzing two different methods for constructing the summary. The first (Sen-Rich) is concerned with selecting summary sentences rich in important topics that balance between relevance and coverage; while the second technique (Doc-Rich) presents a new method for ordering the extracted sentences based on selecting one or more base document. We applied the new summarization techniques in our experiments to summarize different types of Arabic documents.

Both the word representation granularity level, and its extracted morpho-syntactic features directly affect the performance of the keyphrase extraction subsystem and hence the summarizer output. Section 3.1 reviews the Arabic Keyphrase Extractor used, sections 3.2 and 3.3 describe the cluster topics construction, and sentences extraction algorithms.

## 3.1 Local document keyphrase extractor

The first step was to pass each document of the cluster to the keyphrase extractor LBAKE [6] module to extract indicative keyphrases of each document at lemma level. LBAKE is based on three main steps: Linguistic processing; candidate phrase extraction; and feature vector calculation.

LBAKE is a supervised learning system for extracting keyphrases of single Arabic document. The extractor is supplied with linguistic knowledge as well as statistical information to enhance its efficiency. All possible phrases of one, two, or three consecutive words that appear in a given document are generated as n-gram terms. These n-gram words are accepted as a candidate keyphrase if they follow syntactic rules. To hide inflectional variations, words are represented in their lemma forms in all computation processes. The importance of a keyphrase (score) within a free-text document is based on nine features:

1.  Number of words in each phrase.
2.  Frequency of the candidate phrase.
3.  Frequency of the most frequent single word in a candidate phrase.

4.  Location of the phrase sentence within the document.
5.  Location of the candidate phrase within its sentence.
6.  Relative phrase length to its containing sentence.
7.  Assessment of the phrase sentence verb content.
8.  Assessment as to whether the phrase sentence is in the form of a question.

Weights of these features were learned during building the classifier. The output of LBAKE is a set of scored keyphrases normalized to their maximum, representing the input document. Each document is replaced by the features illustrated in Table (1).

Table 1: Features representing a document

| Feature | Description |
| --- | --- |
| $d_i$ | Document number [$1 \leq i \leq D$] |
| $S_{i,j}$ | Set of sentences in document j |
| $Ld_i$ | Length of the document i expressed as the number of sentences in the document. |
| $NP_i$ | Total number of extracted local keyphrases for a document i |
| $P_{i,j}$ | Set of Keyphrases in a document. P is represented in lemma form, $1 \leq i \leq D$, and $1 \leq j \leq NP_i$ |
| $LS_{i,j}$ | Set of Normalized local Keyphrases score in a document. Their ranges: $0 \leq LS \leq 1$, $1 \leq i \leq D$, and $1 \leq j \leq NP_i$<br>$LS_{i,j}$ is the local score divided by a maximum local score of a given document. |

## 3.2. Constructing cluster topics

The next step of the algorithm was to construct the cluster topics $T_k$, and their Cluster scores $TS_k$ for all documents. Extracted local keyphrases have rich information, and can be used in various scoring schemes for cluster topic construction. To realize this process, all local kephrases features of the set of documents were combined together, and each keyphrase was assigned a new global cluster topic score based on its importance on the local document as well as the relevance to all documents of the cluster. The next subsections illustrate steps to construct cluster topics and their scores.

### 3.2.1 Maximum coverage score

A direct solution to construct the cluster topics (T) is to union all local keyphrases.
$$T = \cup P_{i,j} \; 1 \leq i \leq D, \text{ and } 1 \leq j \leq NP_i$$
Since a keyphrase T may appear in many documents, we set the maximum coverage score MCS equal to the maximum local keyphrase score that match T.
$$MCS_k = max(LS_{i,j}), \text{ and } T_k = P_{i,j}$$

Top ranked non-duplicated keyphrases were then selected, which guaranteed the inclusion of all important local keyphrases in the global summary. All important topics in local documents will be included in the summary with this technique, hence it tends to maximize the coverage of the summary.

### 3.2.2 Centroid topic score

In spite of its simplicity, the previous scoring ignores the relevance aspect of selected keyphrases. In multi-document summarization, importance should be given to common information that maintained by many documents. For example if there are two different keyphrases with the same local scores in two different documents in a cluster, and only one of these keyphrases could be repeated multiple times in other documents. To provide a fair assessment of the keyphrase importance, repetitions of the keyphrase in other documents must be considered. This is represented by the relevance feature which reflects the importance of a keyphrase for the set of documents. The relevance of a local keyphrase (P) can be found by its frequency ($F_P$) among the cluster documents. The concept is that the importance of a keyphrase increases as it appears in more documents. The frequency F of a keyphrase P is given by:

$$F_P = count(P \cap P_{i,j}), \quad for\ all\ i,j$$

Note that F also represents the number of documents that contain P. The use of frequency as a sole representation of the importance of a keyphrase is not always an accurate representation of importance. For example a minor topic that is repeated in many documents will gain a false importance. Therefore, we considered a 'Centroid Topic Score' as a solution to overcome this. Centroid topic is defined as the topic that is important in its local document and relevant to document cluster. Therefore, Centroid Topic Score CTS is given by multiplying the two factors:

$$CTS_k = NF_k\ MCS_k \qquad\qquad (1)$$
$$where\ \ NF_{k=} F_k\ /\ max(F)$$

Where $NF_k$ : is the normalized frequency of $T_k$ among T.

### 3.2.3 Centroid document score

An important feature of the proposed summarization system, is the ability to reject (or at least reduce) the effect of non-related documents. For example, if there is a cluster containing nine documents concerned with 'Tsunami', and the tenth article is strongly related to 'terrorist incident'. Since the tenth article is strongly related to 'terrorist incident', its keyphrases still have top scores. This will mislead the extractor to include unimportant topics. In our approach, we exploit a 'Centroid Document Score CDS' to evaluate the relevance of the document to the cluster. Keyphrases extracted from centroid documents get a bonus by CDS values. CDS is ranked by the number of links of a document to other documents. A Link Score between two documents A and B is the count of their matched keyphrases.

CDS of a document k is calculated as the summation of link scores between document k and all other documents divided by the number of keyphrases of k. Since the keyphrases are guaranteed not to be repeated within a local document, CDS is set to:

$$CDS_K = \left( \frac{\sum_{i=1}^{NP_K}(F_i)}{NP_K} \right)$$

In multiple documents, the extracted keyphrase must be important in its local document in addition to having strong relevance to the main concepts of the cluster. Finally, to have a balance between maximum coverage and relevance, we included the centroid document score to equation (1) to represent the Maximum Centroid Topic Score.

$$MaxCR\ TS_K = CDS_K\ CTS_K \qquad\qquad (2)$$

## 3.3. Extraction of MD summary

In this step we presented two techniques to produce relevant summary sentences. Each one of these techniques can achieve one or more of the summarization goals. Production of summary sentences requires two steps. The first is to rank the cluster sentences according to some salient features, and the second is to present these sentences in some order. Both techniques have the ability to produce cluster summaries based on the automatically extracted cluster topics ($T_k$, and their Cluster Scores $TS_k$). The first technique, Sen-Rich, prefers maximum richness sentences along the cluster, while the second, Doc-Rich, prefers sentences from maximum richness documents. In the following subsections, we describe both techniques, and the pros and cons of each.

### 3.3.1 Sentence richness technique (Sen-Rich)

Rich sentences are those that contain many important cluster topics. The basic idea is greedy in the sense that the algorithm tries to select a minimum number of sentences that carry most cluster topics. In the Sen-Rich technique, importance of the sentence is determined by summing all cluster topic scores $TS_k$ that exist in that sentence. All sentences are ranked and top n sentences are selected according to predetermined summary length.

Note that the cluster score of a topic carries both the document importance (relevance), and local topic importance (coverage). Therefore, the algorithm aims to capture sentences that include the important shared common concepts along the cluster, along with the important concepts that are addressed by individual documents.

The Sen-Rich scoring technique tends to be useful when highly condensed summary is required. A few sentences carrying most important topics are selected and presented. However, the algorithm suffers from lack of coherence between the selected sentences, as they belong to different documents. Also, diversity of sentences is not guaranteed, dominant topics (highly scored) may appear in many sentences. Therefore, the presented algorithm is useful for providing highly condensed summary for closely related documents such as news articles.

The summary produced could include similar sentences describe the same subject or explain two important aspects in different ways; the algorithm can avoid (or at least reduces) the appearance of those sentences by accepting a threshold of the Unit Overlap similarity measure (Saggion et al., 2002).

$$overlap(X,Y) = \frac{\| X \cap Y \|}{\| X \| + \| Y \| - \| X \cap Y \|}$$

Where $\| X \cap Y \|$ are the matched keyphrase lemmas between sentences X,Y. ‖X‖ , ‖Y‖ are the total number of keyphrase lemmas in sentences X,Y. The similarity here is measured based on the number of matched keyphrases occurring in the sentences.

### 3.3.2 Document richness technique (Doc-Rich)

For a highly cohesive readable text, we defined a second summarization technique which is based on the centroid document approach discussed in section (3.2.3). A document usually consists of several topics, part of these topics are addressed and maintained by many other documents, while others are individual topics for the document. The basic concept was to construct a method that preferred sentences of centroid documents when more than one sentence carried the same topic. Extracted sentences then follow the context flow of centroid documents, and hence a cohesive summary was produced.

The first step of the algorithm orders documents $D_i$ according to their CDS values. As explained in section (3.2.3) CDS measures the importance of a document in terms of shared concepts with other documents of the cluster. The second step orders cluster topics $T_k$, by cluster scores $TS_k$. Then for each $T_k$, the algorithm searches ordered $D_i$ for the first sentence that include $T_k$. When found, the sentence is appended to summary sentences if it does not already exist in the list. The process is continued until all sentences representing cluster topics are extracted. The algorithm then selects first n % number of sentences required for summary.
The Doc-Rich algorithm avoids the errors that occurred as a result of repeated documents with different names, or two sentences describing the same subject. As it is not allowed for the cluster topics $T_k$ to have a duplicate keyphrase, and this thereby avoids duplicate sentences (or even sentences that include same important keyphrases), since for each keyphrase, only one sentence is extracted.

# 4. EVALUATION OF THE PROPOSED TECHNIQUES

Evaluation of automatic text summarization systems by human evaluators requires massive efforts. This hard expensive effort has held up researchers looking for methods to evaluate summaries automatically. Evaluating summarization systems is not a straightforward process since it is an elusive property [9]. Current automated methods compare fragments of the summary to be assessed against one or more reference summaries (typically produced by humans), measuring how much fragments in the reference summary is present in the generated summary. In order to evaluate the performance of the proposed summarization techniques, two experiments were carried out. The first experiment was designed to evaluate and compare the accuracy of the proposed system against other systems using automatic measure. The second was designed to evaluate other features of the proposed summarization techniques given different article types. Human evaluators were used in this experiment. The following subsections describe the details of each experiment and the datasets used.

## 4.1 The datasets

To test the new keyphrase-based techniques, two datasets were adopted. The first (DataSet1) was TAC 2011 Dataset. It was selected to evaluate and compare the summaries of the new techniques against the published summaries of systems presented in TAC 2011, using the same dataset for the same task of producing 240-250 words summary. DataSet1 included the source texts, system summaries, and human summaries. The data set is available in 7 languages including Arabic (http://www.nist.gov/tac/2011/ Summarization/). It was derived from publicly available WikiNews English texts. Texts in other languages have been translated by native speakers of each language. The source texts contain ten collections of related newswire and newspaper articles. Each collection contains a cluster of ten related articles. The average number of words per article is 235 words. Each cluster of related articles deals with news about a single event. Each article in a certain cluster includes a limited number of topics (mostly one or two topics) about the event. Some topics are dealt with by many documents, while others are addressed in a single document.
DataSet1 is not enough to test all the features of the proposed algorithms, since all documents are concerned with a single event that carries a limited number of topics. Therefore, we then collected a second dataset (DatSet2) that contains four collections of related web articles in social, science, geology, and geophysics domains. Each collection contains a cluster of 10 related articles with multiple numbers of topics for each domain subject. The average number of words per article is (340) words. The summarization task was to produce a 280-290 word summary.

## 4.2 Experiment 1: System validation against other systems

Many systems have been developed for automatic evaluation of summary systems. One such system, Recall-Oriented Understudy for Gisting Evaluation (ROUGE) [14], is a recall measure that counts the number of overlapping n-gram units between the system summary generated by computer and several reference summaries produced by human. ROUGE has proved to be a successful algorithm. Several variants of the measure were introduced, such as ROUGE-N, and ROUGE-S. ROUGE-N is an n-gram recall between a candidate summary and a set of reference summaries. ROUGE-S utilizes Skip-Bigram Co-Occurrence Statistics. Skip-bigram any pair of words in their sentence order, allowing for arbitrary gaps. ROUGE-S, measures the overlap ratio of skip-bigrams between a candidate summary and a set of reference summaries. DataSet1 described in sec (4.1) was used for this experiment.

This experiment has been applied to measure the success of the proposed summarization techniques by comparing them with other systems. To carry out the experiment, the two proposed techniques were applied to summarize the same cluster sets used in TAC 2011 to produce 240-250 word summaries for each cluster. DataSet1 already includes the summarization results generated by other systems under test as well as the reference human summaries. Rouge test was applied to all summaries to be compared, and reported the experimental results in terms of ROUGE-2 and ROUGE-S. Precision P, recall R, and F-measure were calculated for each measure. We have used the existing Rouge evaluation methods implemented in the Dragon Toolkit Java based package (http://dragon.ischool.drexel.edu/ ). The existing Arabic lemmatizer [5] was used in stemming for all summaries.

Tables (2, 3) show a comparison between the two new techniques (Sen-Rich, and Doc-Rich) among different summarization systems in terms of ROUGE-2 and ROUGE-S measures. The results show that the two proposed techniques perform well compared to other systems. Sen-Rich technique outperformed all systems in ROUGE-S measure; it gains 0.2014, compared to 0.1585 for Classy and 0.1554 for Doc-Rich.

The positive results for the two proposed techniques, particularly Sen-Rich technique, were expected. The good performance of the two techniques lies in the initial usage of a good scoring scheme for cluster topics, and then adopting these topics as attributes to evaluate the topic's richness of sentences and documents within the cluster, instead of using text passages (documents, paragraphs, or sentences) directly for assessment. In Sen-Rich technique, the scoring algorithm is based on summing for all cluster topic scores existing in that sentence. The algorithm captures rich sentences that contain many important cluster topics. For a condensed summary, and the cluster of documents dealing with a single event with a limited number of topics, the algorithm succeeds to capture a minimum number of sentences that carry the most important topics of the cluster. Doc-Rich technique ranked a satisfactory result, but less than Sen-Rich.

In Doc-Rich technique, the algorithm extracts only one sentence for each keyphrase. It starts by high score keyphrases, then extracts the first seen sentence in centroid document that contains this keyphrase for summary; if it is not already existing. In this technique, only one sentence at most is extracted for each keyphrase. This gives the opportunity to cover all the major topics of the document, and at the same time, allows for keeping redundancy to a minimum. We expect Doc-Rich to function with consistently good coverage when summarizing multiple topic documents. Moreover, in this technique, the algorithm prefers sentences of centroid document, extracted sentences then follow the context flow of this document, and hence a cohesive summary is produced. Automatic evaluation systems do not deal with the coherence feature in its evaluation. We therefore performed a second experiment, using human evaluators to measure the performance of the two proposed techniques when applied to summarize multiple topic documents.

Table 2: A comparison between the proposed techniques and different systems using ROUGE-2

| System | F-Measure |
|---|---|
| CLASSY1 | 0.1529 |
| **Sen-Rich** | **0.1511** |
| **Doc-Rich** | **0.1432** |
| TALN_UPF1 | 0.1326 |
| CIST1 | 0.1279 |
| UoEssex1 | 0.1165 |
| UBSummarizer1 | 0.0915 |

Table 3. A comparison between the proposed techniques and different systems using ROUGE-S

| System | F-Measure |
|---|---|
| **Sen-Rich** | **0.2014** |
| CLASSY1 | 0.1585 |
| **Doc-Rich** | **0.1554** |
| TALN_UPF1 | 0.1472 |
| UoEssex1 | 0.1331 |
| CIST1 | 0.0973 |
| UBSummarizer1 | 0.0838 |

## 4.3 Experiment 2: Multiple topics test

In this experiment we compared the two techniques to determine their ideal applicability to extract summary sentences given different article types. In the first experiment the documents to be summarized are concerned with a single event that carries a limited number of topics. However, in this experiment, we applied the proposed techniques to summarize clusters of multi-topic articles using DataSet2 described in section (4.1). Three human evaluators were asked to assess the resulting summaries. Summaries were evaluated on the bases of measuring their coverage, informative richness and coherence. The maximum score was 5 per measure with a total 15 for a summary. Figure 1 is an example of summary output by Doc-Rich technique.

Table 4 shows the three human evaluation scores (H1:H3) for the four clusters (C1:C4) for the two proposed techniques. It is noticed in this test that Doc-Rich technique performed better than Sen-Rich technique. Doc-Rich had an average score of 11, while Sen-Rich had an average score of 9.2. The Doc-Rich technique was extracted the summary sentences as much as possible from a centroid document to produce a cohesive readable text summary. In this technique documents were sorted according to their importance. Then the algorithm extracted the first seen sentence in the list of documents. Most of the summary sentences were extracted from the centroid (top) document. In our experiment, we found that on the average, 72 % of the summary sentences were extracted from the centroid document. Ordering the extracted sentences was provided by the document itself, and hence a cohesive summary was produced.

In Sen-Rich technique, many sentences that describe same (focus) topics dominate the selection. All sentences containing the main topics will get higher scores, which may lead to redundant sentences selection, and does not provide room for other topics to appear in the summary. However, the Doc-Rich technique extracts only one sentence for each important cluster topic, this provides an opportunity for more important concepts to appear in summary, and hence more coverage.

Table 4: Human evaluations scores for the proposed techniques using DataSet2

| | | H1 | H2 | H3 |
|---|---|---|---|---|
| Doc-Rich | C1 | 11 | 12 | 11 |
| | C2 | 12 | 13 | 11 |
| | C3 | 10 | 12 | 10 |
| | C4 | 8 | 10 | 12 |
| Sen-Rich | C1 | 7 | 10 | 11 |
| | C2 | 10 | 10 | 9 |
| | C3 | 9 | 11 | 9 |
| | C4 | 7 | 8 | 9 |

خدمات الشبكات الاجتماعية
تقدم الشبكات الاجتماعية أو صفحات الويب خدمات عديدة لمتصفحيها
فهي تتيح لهم حرية الاختيار لمن يريدون في المشاركة معهم في اهتماماتهم
وبظهور شبكات التواصل الإجتماعي مثل الفيس بوك و غيره
توسعت الخدمات المرجوة من هذه الشبكات ومنحت متصفحيها إمكانيات واسعة في تبادل المعلومات في مجالات التعليم والثقافة والرياضة وغيرها
وهذه الشبكات هي عبارة عن مواقع إلكترونية إجتماعية
"وهي مواقع انتشرت في السنوات الأخيرة بشكل كبير وأصبحت أكبر وأضخم مواقع في فضاء الويب ولازالت مستمرة في الانتشار الأفقي المتسارع
حيث يمكن لأحد المستخدمين الارتباط بأحد الأصدقاء عبر الموقع ليصل جديد ما يكتب ويضيف ذلك الصديق إلى صفحة صديقه
ومن الخدمات التي تقدمها هذه الشبكات هي: إتاحة المجال للأفراد في الدخول إلى المواقع الإجتماعية والتعريف بأنفسهم ومن ثم التواصل مع الأخرين الذين تربطهم بهم اهتمامات مشتركة
وتنقسم المواقع الإجتماعية إلى قسمين رئيسين هما: القسم الأول: هي مواقع تضم أفراد أو مجاميع من الناس تربطهم إطارات مهنية أو إجتماعية محددة
القسم الثاني: هي مواقع التواصل الإجتماعي المفتوحة للجميع ويحق لمن لديه حساب على الإنترنت الانضمام إليها واختيار أصدقائه والتشبيك معهم وتبادل الملفات والصور ومقاطع الفيديو وغيرها
أو التي تسمى شبكات التواصل الإجتماعي على الإنترنت
وإنها الركيزة الأساسية للإعلام الجديد أو البديل
فقد أقبل عليها ما يزيد عن ثلثي مستخدمي شبكة الإنترنت
تقدم شبكات التواصل الإجتماعي خدمات لمستخدميها ممن لديهم اهتمامات متشابهه سواء أكانوا زملاء دراسة أو عمل أو أصدقاء جدد
ويعتبر موقع MySpace من أوائل وأكبر الشبكات الاجتماعية على مستوى العالم ومعه منافسه الشهير فيس بوك الذي

Figure 1: Example summary output by Doc-Rich technique

## 5. CONCLUSIONS

In this research we have presented two keyphrase-based techniques for multi-document summarization. A new concept-centroid approach was presented. In both techniques a centroid cluster topics scoring scheme was used to recognize the importance of a particular keyphrase. In the first technique Sen-Rich, sentence importance was determined by summing all cluster topic scores that exist in that sentence. In the second technique, Doc-Rich, documents were sorted according to their importance; the most important one was the centroid document. For each important cluster topic only one sentence was extracted from centroid document. We conducted two experiments. In the first, we compared the proposed techniques with different systems presented at TAC 2011 to summarize clusters of Arabic documents. We employed ROUGE evaluation measures. The results show that Sen-Rich technique had superiority over all systems in ROUGE-S. In the second experiment, clusters of multiple topic web articles were collected. Using human evaluators, the results showed that Doc-Rich technique had superiority over Sen-Rich technique. The two experiments show that Sen-Rich technique tends to be useful when the documents to be summarized are dealing with single event with limited number of topics (e.g news articles) and there is a highly condensed summary. The algorithm succeeds to capture sentences that carry the most important topics of the cluster. However, for a task of summarizing multiple documents with multiple numbers of topics, Doc-Rich technique tends to be more appropriate for better coverage and a cohesive readable text summary.

# REFERENCES

[1] Barzilay, R. (2003), Information Fusion for MultiDocument Summarization: Paraphrasing and Generation, PhD Thesis, Columbia University.

[2] Barzilay, R., & Elhadad, M. (1997), Using lexical chains for text summarization, In Proceeding of the Intelligent Scalable Text Summarization Workshop (ISTS'97), ACL.

[3] Bollegala, D. T. (2007), Improving coherence in multi-document summarization through proper ordering of sentences, Master of Science, Information & Communication Engineering, Graduate School of Information Science and Technology,Tokyo, Japan.

[4] El-Haj, M., Kruschwitz, Udo., Fox, C. (2011), Exploring Clustring for Multi document Arabic Summarization, Information Retrieval Technology - 7th Asia Information Retrieval Societies Conference, AIRS 2011, Dubai, United Arab Emirates.

[5] El-Shishtawy, T. & El-Ghannam, F. (2012), An Accurate Arabic Root-Based Lemmatizer for Information Retrieval Purposes, International Journal of Computer Science Issues, Volume 9, Issue 1.

[6] El-Shishtawy, T. & El-Ghannam, F. (2012), Keyphrase Based Arabic Summarizer (KPAS), 8th International Conference on Informatics and Systems (INFOS).

[7] Giannakopoulos, G., El-Haj, M., Favre, B., Litvak, M., Steinberger, J., & Varma, V. (2011), TAC2011 MultiLing Pilot Overview, TAC 2011 Workshop, Gaithersburg, MD, U.S.A.

[8] Goldstein, J., Mittal, V., Carbonell, J., and Kantrowitzt M. (2000), Multi-Document Summarization By Sentence Extraction. NAACL-ANLP 2000 Workshop on Automatic summarization, Vol 4.

[9] Hassel. (2004), Evaluation of Automatic Text Summarization - A practical implementation. Licentiate thesis, Department of Numerical Analysis and Computer Science, Royal Institute of Technology, Stockholm, Sweden.

[10] Hatzivassiloglou, V., Klavans, J., Holcombe, L., Barzilay, R., Kan, M., and Kathleen, R. McKeown, SimFinder, (2001), A Flexible Clustering Tool for Summarization. NAACL Workshop on Auromatic Summarization, Association for Computational Linguistics.

[11] Inderjeet and Mark, T. Maybury, (2001), Advances in automatic text summarization, The MIT Press.

[12] Ledeneva, Y., García-Hernández, R.,Vazquez-Ferreyra, A., de Jesús, N. (2011). Experimenting with Maximal Frequent Sequences for Multi-Document Summarization. In Text Analysis Conference TAC 2011, USA.

[13] Lin, C.Y. and Hovy, E., (2001), Neats:a multidocument summarizer, Proceedings of the Document Understanding Workshop (DUC).

[14] Lin, C. Y., (2004), Rouge: A package for automatic evaluation of summaries". Proceedings of the Workshop on Text Summarization Branches Out (WAS 2004), pages 25–26.

[15] Liu H., Zhao Q., Xiong Y., Li L., Yuan, C., (2011), The CIST Summarization System at TAC 2011, In Text Analysis Conference TAC 2011, USA.

[16] Okazaki, N., Matsuo, Y., and Ishizuka, M., (2004), Improving chronological sentence ordering by precedence relation. In Proceedings of 20th International Conference on Computational Linguistics (COLING 04), pages 750–756.

[17] Radev, Dragomir R., Otterbacher, J., Qi, H., and Tam, D., (2003), Mead reducs: Michigan at duc 2003. In Proceedings of DUC 2003. Edmonton, AB, Canada.

[18] Radev, Dragomir R., J, H., Stys, M., and Tam, D. , (2004), Centroid-based summarization of multiple documents. Information Processing and Management. in press.

[19] Radev, D., Otterbacher, J., Winkel, A., & Blair-Goldensohn, S., (2005), NewsInEssence: summarizing online news topics. Communications Communications of the ACM, 48(10), 95-98.

[20] Saggion H., (2011), Using SUMMA for Language Independent Summarization at TAC 2011. In Text Analysis Conference TAC 2011, USA.

[21] Saggion, H., Radev, D., Teufel, S., Lam, W., Strassel, S., (2002), Developing Infrastructure for the Evaluation of Single and Multi-Document Summarization Systems in a Cross-Lingual Environment, In Proceedings of LREC, Las Palmas, Spain.

[22] Salton G. and McGill M. J. (1988), Introduction to Modern Information Retreival. McGraw-Hill.

[23] Schlesinger J., D., O'Leary, D., P., and Conroy, J., M., (2008), Arabic/English multi-document summarization with classy: the past and the future, In Proceedings of the 9th international conference on Computational linguistics and intelligent text processing, CICLing'08, Berlin. pages 568–581.

## Authors

**Fatma El-Ghannam** is a Researcher Assistance at Electronics Research Institute – Cairo, Egypt. She has significant research interests in Arabic language generation and analysis. Currently, she is preparing for a Ph.D. degree in NLP.

**Tarek El-Shishtawy** is a Professor at Faculty of computers and information, Benha University- Benha, Egypt. He participated in many Arabic computational Linguistic projects. He paritipated in Large Scale Arabic, annotated Corpus, 1995, which was one of important projects for Egyptian Computer Society, and Academy of Scientific Research and Technology. He has many publications in Arabic Corpus, machine translation, Text, and data Mining.